\pdfoutput=1
\documentclass[11pt]{article}

\usepackage[T1]{fontenc}
\usepackage[utf8]{inputenc}
\usepackage{lmodern}
\usepackage{amsmath,amssymb}
\usepackage{graphicx}
\usepackage{subcaption}
\usepackage{booktabs,array,tabularx}
\usepackage[margin=1in]{geometry}
\usepackage{xcolor}
\usepackage{microtype}
\usepackage{etoolbox}
\usepackage{url}
\usepackage[numbers,sort&compress]{natbib}
\usepackage[hidelinks]{hyperref}

\newcommand{\backmatter}{%
  \clearpage
  \setcounter{secnumdepth}{0}%
}

\title{\textbf{Consolidator: Learning Persistent Routed Memory Across Context Boundaries}}
\author{%
  Sungwoo Goo$^{1}$ \qquad Hwi-yeol Yun$^{1}$ \qquad Sangkeun Jung$^{2}$\\
  \small $^{1}$College of Pharmacy, Chungnam National University\\
  \small $^{2}$Department of Computer Science \& Engineering, Chungnam National University\\
  \small Daejeon, Republic of Korea\\
  \small \texttt{swgoo91@gmail.com, hyyun@cnu.ac.kr, hugmanskj@gmail.com}%
}
\date{}

\begin{document}
\maketitle

\begin{abstract}
Copying short-term memory (STM) into a slower store can preserve state across a context boundary, but persistence alone does not ensure that the retained state influences subsequent memory access.
We test this distinction in a Phasor Memory Network (PMNet) using Consolidator, a shared slot-local operator that transforms routed STM before accumulating it into long-term memory (LTM), without replaying the source tokens.
After each consolidation, the KV cache and STM are cleared.
The retained LTM can still be read and is also fed into the hierarchical router, thereby conditioning which explicit-memory slots subsequent inputs access.
We evaluate this mechanism on a two-segment modulo-10 mapping task in which the second segment updates the mapping at the same memory address.
Following a second consolidation and reset, a held-out query must recover the updated mapping from LTM.
The backbone and memory interface are frozen, leaving only 12.35K Consolidator parameters trainable (0.041\% of a 29.95M model).
Across five paired runs from the same STM-pretraining checkpoint, direct LTM routing raises updated-mapping recall from $44.38\pm1.94\%$ to $87.02\pm1.76\%$ ($+42.64\pm1.10$ percentage points), while immediate STM recall remains 89.90\% in both conditions; both train separate Consolidators and retain the same LTM read paths.
Learned consolidation outperforms forced identity accumulation by $21.40\pm1.91$ percentage points without routing and $68.70\pm1.76$ with routing.
Thus, on this task, consolidated LTM serves as both retrievable content and an access state that shapes subsequent slot selection.
\end{abstract}


\section{Introduction}\label{1-introduction}

Transformer context is largely an append-only record of token and KV history \citep{vaswani2017attention}.
A writable memory offers a different abstraction: experience can be compressed into bounded state, updated at an address, and retrieved after transient context is discarded.
Such a system must learn to form useful short-term state, preserve or revise it across boundaries, and use the retained state to select memory slots for later writes and reads.

Merely detaching and copying STM into a slower buffer would provide carryover, but it would leave two mechanistic questions unresolved.
Does a fixed pretrained memory interface require a learned transition to revise conflicting content, rather than raw accumulation?
Does the retained state only supply values to the read path, or does the router also use it to select slots for subsequent writes and reads?

Differentiable memory, segment recurrence, compression, and adaptive state demonstrate several parts of this lifecycle \citep{graves2014ntm,graves2016dnc,dai2019transformerxl,rae2019compressive,sun2024ttt,behrouz2024titans}.
We ask two linked questions: \textbf{can a separately trained operator convert useful routed STM into LTM, and can that LTM guide later slot selection after the KV cache and STM are cleared?}

We study this question in PMNet, where tokens write phase-valued state through a hierarchical router and memory is separable from the local attention cache \citep{goo2026pmnet}.
Consolidator applies one shared gated phase transform to occupied STM slots, accumulates the result into LTM, and then permits KV and STM to be cleared.
In later segments, the model can still read the retained LTM.
The router also includes that LTM state in the phase-valued representation used to score slots at the same hierarchy level.
As a result, information first written to STM can influence which memory slots later inputs select.

The controlled task comprises two context segments that reuse the same address and rule family, while the second segment replaces the first segment's function parameters.
The final held-out query targets this updated mapping and is answered only after both demonstration contexts have been processed and the KV cache and STM have been cleared.
In the central intervention, 99.959\% of the model is frozen and only the 12.35K-parameter Consolidator is trained.
A paired routing ablation retains learned consolidation and the LTM read path but removes the direct LTM input to same-level slot routing.
This reduces updated-mapping LTM recall from \(87.02\pm1.76\%\) to \(44.38\pm1.94\%\), a paired decrease of \(42.64\pm1.10\) pp, while immediate STM recall of the second mapping remains exactly 89.90\% in both conditions.
The ablation therefore tests whether retained LTM guides later slot selection in addition to supplying retrievable content.
A complementary identity intervention tests learned revision against raw STM accumulation: the learned-minus-identity gap is \(21.40\pm1.91\) pp without direct routing and \(68.70\pm1.76\) pp with routing, while mismatched and fresh LTM controls remain near chance.
A dual-objective experiment tests whether pre-consolidation STM recall can coexist with post-reset LTM recall.

Our contributions are:

\begin{itemize}
\item
  A shared slot-local transform that consolidates routed latent STM without replay or topology-dependent parameter growth.
\item
  Direct LTM-conditioned slot routing as an architectural inductive bias that lets retained state guide later writes and reads through a frozen router.
\item
  A sequential same-address update task that separates four memory functions: carrying state across a reset, revising an existing memory, retrieving retained content, and using LTM to guide slot selection.
  Same-checkpoint, content-replacement, and routing interventions isolate these functions.
\end{itemize}

The evidence is a mechanism-level proof of concept, not yet a claim about natural-language long context, continual learning, or systems efficiency.

\section{Related Work}\label{2-related-work}

\subsection{Explicit and recurrent memory}\label{21-explicit-and-recurrent-memory}

Neural Turing Machines and Differentiable Neural Computers established end-to-end learned addressing over external read-write memory \citep{graves2014ntm,graves2016dnc}.
Transformer-XL and Compressive Transformer carry or compress activations across segments \citep{dai2019transformerxl,rae2019compressive}; Recurrent Memory Transformer and Infini-attention maintain bounded recurrent state \citep{bulatov2022rmt,munkhdalai2024infini}; and Memorizing Transformers retrieve earlier representations from a non-differentiable store \citep{wu2022memorizing}.
These systems preserve or retrieve history, whereas our intervention asks a learned slot-local transition to revise conflicting state at the same routed address and then feed the consolidated result directly into subsequent slot routing.

PMNet is the architectural base of this study.
It represents recurrent memory updates as phasor rotations and organizes addresses hierarchically \citep{goo2026pmnet}.
We add an explicit STM--LTM boundary and isolate its function rather than revisiting PMNet's language-modeling or copy benchmarks.

\subsection{Adaptive state and compact adaptation}\label{22-adaptive-state-and-compact-adaptation}

Fast weights, selective state-space models, Test-Time Training layers, and Titans all make computation depend on rapidly changing latent state \citep{ba2016fast,gu2023mamba,sun2024ttt,behrouz2024titans}.
Unlike test-time gradient methods, Consolidator keeps model parameters fixed during inference.
Its forward pass updates non-parametric memory, which then guides later slot selection.
Context Distillation instead stores and routes independent LoRA parameter memories \citep{zheng2026context}; PMNet writes newly observed content directly into routed non-parametric memory.

Textual Inversion showed that a frozen model can use a small learned embedding to represent a new concept \citep{gal2023textual}.
One-layer post-training likewise found that adapting a restricted Transformer layer can recover much of full-parameter improvement \citep{zhang2026onelayer}.
Both rely on gradient optimization and therefore do not demonstrate forward-only acquisition during a memory episode.
They support a more limited capacity premise: pretrained computation can make effective use of a compact adaptation substrate.
PMNet trains the memory interface and Consolidator end to end; after training, new content is written and consolidated without changing model parameters.

\subsection{Memory consolidation}\label{23-memory-consolidation}

Complementary Learning Systems motivates distinct fast and slow stores, while artificial consolidation commonly combats forgetting through stored or generated replay \citep{mcclelland1995cls,hayes2021replay}.
Our terminology is an engineering analogy, not a claim of biological equivalence.
Auto-Dreamer rewrites symbolic agent memory from stored entries and trajectories \citep{ye2026autodreamer}.
At each boundary, Consolidator transforms the routed STM directly into LTM without revisiting the input sequence.
That LTM is later both retrieved as stored content and used to determine which slot new inputs access at the same memory level.

\section{Problem Formulation and PMNet Background}\label{3-problem-formulation-and-pmnet-background}

\subsection{Memory lifecycle}\label{31-memory-lifecycle}

\begin{figure}[t]
  \centering
  \begin{subfigure}[t]{0.23\textwidth}
    \centering
    \includegraphics[width=\linewidth]{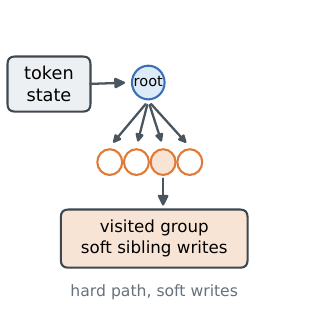}
    \caption{Routed writable STM.}
  \end{subfigure}\hfill
  \begin{subfigure}[t]{0.30\textwidth}
    \centering
    \includegraphics[width=\linewidth]{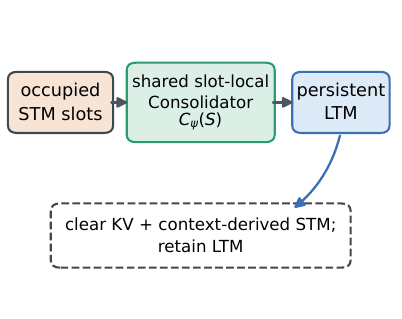}
    \caption{Boundary consolidation.}
  \end{subfigure}\hfill
  \begin{subfigure}[t]{0.45\textwidth}
    \centering
    \includegraphics[width=\linewidth]{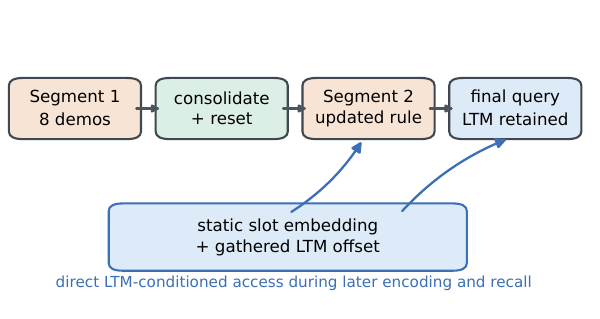}
    \caption{Two-segment memory episode and routed reuse.}
  \end{subfigure}
  \caption{\textbf{Routed latent-memory lifecycle.}
  Tokens write routed STM; at a boundary, Consolidator updates persistent LTM and KV/STM are cleared.
  Retained LTM supports later reads and conditions subsequent explicit-memory slot selection.}
  \label{fig:lifecycle}
\end{figure}

A memory episode comprises segments $X_1,\ldots,X_T$ and three non-parametric states: a local sliding-attention KV cache $K_t$, routed short-term memory $S_t$ written within segment $t$, and long-term memory $L_t$ retained across segments.
We distinguish two functions of retained LTM: \emph{content state} supplied to the read path and \emph{access state} that conditions which explicit-memory slots subsequent inputs select.
Our evaluation asks four questions: whether STM contains the segment-specific mapping before consolidation; whether that mapping can be recovered from LTM after the KV cache and STM are cleared; whether learned consolidation can replace the mapping already stored at a reused address; and whether supplying LTM directly to the router changes post-reset recall relative to making it available only through the read path.

Copying a detached STM snapshot into LTM would alter training credit assignment and preserve state across the reset, but persistence alone would not establish either learned revision or direct control over subsequent slot selection.
We therefore use two controlled comparisons.
Comparing learned consolidation with identity accumulation tests whether a learned transformation at the STM--LTM boundary is needed to revise retained content.
Comparing direct LTM routing on and off tests whether consolidated LTM improves recall by guiding subsequent slot selection, beyond supplying stored content through the read path.

\subsection{Hierarchical routed writes}\label{32-hierarchical-routed-writes}

PMNet represents memory dimensions as phase angles.
At hierarchy block $b$, a token visits group $g$ with candidate child slots $j$.
Before direct LTM conditioning is introduced, the phase-valued representation used to score each candidate slot is
\begin{equation}\label{eq:base-routing-state}
a_{t,b,g,j}=u_{t,b,g,j}+e_{b,g,j},
\end{equation}
where \(u\) is incoming latent state and \(e\) is a learned static slot embedding.
With $\phi(a)=[\sin(a);\cos(a)]$, token and slot projections define a cosine-similarity distribution $p_{t,j}$ over siblings.
The highest-scoring child selects the next group, while every sibling receives a differentiable phase update
\begin{equation}\label{eq:soft-stm-write}
\Delta S_{t,j}=p_{t,j}\,\pi\tanh\!\left(
W_oW_v\operatorname{RMSNorm}(h_t)\right).
\end{equation}
Thus traversal is hard top-1 between levels but writes remain soft within each visited group.
An occupancy mask records the groups to consolidate.
Reads use the wrapped dynamic state
\begin{equation}\label{eq:memory-read-state}
M_{b,g,j}=(S_{b,g,j}+L_{b,g,j}) \mod 2\pi,
\end{equation}
together with the static embeddings.
Further PMNet details are given by \citet{goo2026pmnet}.

\subsection{Operational meaning of replay-free}\label{33-operational-meaning-of-replay-free}

We call the consolidation procedure replay-free because each demonstration segment is presented only once along the main sequential trajectory that forms and updates LTM.
At each boundary, Consolidator receives only routed STM and an occupancy mask; it neither re-encodes the original demonstration tokens nor retrieves them from an episodic replay buffer.
After the reset, the final query is answered using retained LTM without presenting the demonstrations again.
The dual-objective auxiliary trajectory separately reprocesses each segment's context to measure pre-consolidation STM recall, but it does not alter the replay-free trajectory used to construct LTM.

Replay-free does not mean BPTT-free.
The present two-segment experiment retains the differentiable graph across both consolidation boundaries.
Detached or truncated training over longer memory episodes remains untested.

\section{Learned Latent-State Consolidation}\label{4-learned-latent-state-consolidation}

\subsection{Shared slot-local phase transform}\label{41-shared-slot-local-phase-transform}

For an occupied STM slot $S\in\mathbb{R}^{d_m}$, define
\begin{equation}\label{eq:phasor-embedding}
z(S)=[\cos S;\sin S].
\end{equation}
A gated MLP shared across all blocks, groups, and slots produces
\begin{equation}\label{eq:consolidator-transform}
\begin{aligned}
r_\psi(z)
&=W_d\left[\operatorname{SiLU}(W_gz)\odot W_uz\right]+b_d,\\
\begin{bmatrix}c_\psi(S)\\s_\psi(S)\end{bmatrix}
&=z(S)\otimes r_\psi(z(S)),\\
C_\psi(S)
&=\operatorname{atan2}\!\left(s_\psi(S),c_\psi(S)\right).
\end{aligned}
\end{equation}
where \(\otimes\) is element-wise complex multiplication in paired cosine/sine coordinates and \(\operatorname{atan2}\) is applied element-wise.
The experiments use \(d_m=32\) and hidden dimension \(d_c=64\).
Zero output weights and unit-phasor bias make \(C_\psi(S)=S\) in Equation~\eqref{eq:consolidator-transform}, so learning begins from exact identity.
The transform is slot local and shares 12.35K parameters regardless of tree capacity.

\subsection{Persistent accumulation and reset}\label{42-persistent-accumulation-and-reset}

For hierarchy block \(b\), group \(g\), and child slot \(j\), let \(S_{b,g,j},L_{b,g,j}\in\mathbb{R}^{d_m}\) denote the STM and LTM phase vectors immediately before consolidation, and let \(L^+_{b,g,j}\) denote the LTM phase vector afterward.
The binary mask \(O_{b,g}\) indicates whether group \(g\) received an STM write during the current segment.
The boundary update is
\begin{equation}\label{eq:ltm-boundary-update}
L^+_{b,g,j}=
\begin{cases}
\!\left(L_{b,g,j}+C_\psi(S_{b,g,j})\right) \mod 2\pi,
& O_{b,g}=1,\\
L_{b,g,j}, & O_{b,g}=0.
\end{cases}
\end{equation}
If no LTM has previously been stored, \(L_{b,g,j}\) is initialized to the zero phase vector.
STM, occupancy, and KV state are then cleared while LTM remains.
The identity control replaces \(C_\psi(S_{b,g,j})\) in Equation~\eqref{eq:ltm-boundary-update} with raw \(S_{b,g,j}\); its forward update is therefore raw copy-and-accumulate, subject only to phase wrapping.
Both modes accumulate rather than overwrite state.

\subsection{Direct LTM-conditioned routing}\label{43-direct-ltm-conditioned-routing}

For a subsequent segment, our extension augments the base routing state in Equation~\eqref{eq:base-routing-state} with the LTM state of the currently visited group:
\begin{equation}\label{eq:ltm-conditioned-routing}
a_{t,b,g,j}=u_{t,b,g,j}+e_{b,g,j}+L_{b,g,j}.
\end{equation}
The static embedding remains a parametric address anchor, while LTM becomes an experience-dependent offset.
Consolidator receives no route label; task loss trains its output through the existing router and read path.
Combining Equations~\eqref{eq:ltm-boundary-update} and~\eqref{eq:ltm-conditioned-routing} yields the recurrent path
\begin{equation}\label{eq:recurrent-memory-path}
S_t \xrightarrow{C_\psi} L_t
\longrightarrow a_{t+1}
\longrightarrow S_{t+1},
\end{equation}
so a fixed router can make experience-dependent slot selections because its non-parametric LTM input changes.
This path makes LTM an access state rather than only retrievable content.

STM accumulated during a segment is not added directly to the candidate-slot representation used by the router at the same hierarchy level.
It can nevertheless influence routing indirectly: reads from earlier hierarchy levels alter the hidden states received by deeper routers.
Because LTM remains fixed within a segment, excluding direct same-level STM feedback avoids a token-to-token routing dependency and preserves parallel routing and write aggregation across the segment.
Consequently, the routing ablation removes only the direct \(L_{b,g,j}\) term in Equation~\eqref{eq:ltm-conditioned-routing}: the router itself, LTM retrieval, and indirect influence on deeper levels remain active.

\subsection{Objectives}\label{44-objectives}

The primary consolidation experiments optimize only the post-reset query for the updated mapping in the second segment:
\begin{equation}\label{eq:updated-objective}
\mathcal{L}_{\mathrm{updated}}
=\operatorname{CE}(f(q_2;L_2),y_2).
\end{equation}
Their checkpoints are selected by validation recall of this updated mapping.

In a separate dual-objective experiment, we test whether one parameter set can support both immediate STM recall and post-reset LTM recall.
For each of the two segments, an auxiliary trajectory evaluates its query after the demonstration has formed STM but before consolidation; \(\mathcal{L}_{\mathrm{STM}}\) is the mean of these two query losses.
Adding this term to Equation~\eqref{eq:updated-objective} gives
\begin{equation}\label{eq:dual-objective}
\mathcal{L}_{\mathrm{dual}}
=\mathcal{L}_{\mathrm{updated}}+\mathcal{L}_{\mathrm{STM}},
\end{equation}
and selects checkpoints by the mean of updated-mapping LTM recall and pre-consolidation STM recall.

\section{Controlled Sequential Same-Address Update Task}\label{5-controlled-same-address-rule-update-task}

\subsection{Procedural memory episodes}\label{51-procedural-memory-episodes}

Each memory episode contains two context segments and one active memory address selected from four address tokens.
One rule family, ADD10 or AFFINE10, is sampled per episode; the first and second segments use different function parameters from that family.
Parameters are resampled across episodes, preventing a fixed address-to-rule solution.

\begin{table}[t]
\centering
\caption{\textbf{Procedural rule families.} Function parameters are resampled for each memory episode.}
\label{tab:task-rule-families}
\small
\begin{tabular}{@{}lll@{}}
\toprule
Family & Function & Sampled parameters \\
\midrule
ADD10 & \(y=(x+k)\bmod 10\) & \(k\in\{1,\ldots,9\}\) \\
AFFINE10 & \(y=(ax+b)\bmod 10\) & \(a\in\{2,\ldots,9\}\), \(b\in\{0,\ldots,9\}\) \\
\bottomrule
\end{tabular}
\end{table}

Each segment provides eight demonstrations and one held-out query, each encoded as
\begin{verbatim}
[address, rule-family, input x, delimiter, answer y]
\end{verbatim}
Loss is applied only to the answer token.
Demonstration and query inputs are distinct within each segment, and the final query is selected so that the first and second mappings give different answers.
The final prediction therefore cannot receive credit for copying an observed answer or retaining only the stale rule.

\subsection{Two-stage training and evaluation protocol}\label{52-two-stage-protocol}

\textbf{STM-pretraining stage (Phase 1)} trains same-segment rule induction from routed STM.
The model processes demonstrations, clears KV history, and answers the held-out query; Consolidator is frozen.
All main consolidation-training conditions share the resulting STM-capable checkpoint.

\textbf{Consolidation-training stage (Phase 2)} processes the two demonstration segments sequentially, consolidates STM into LTM after each segment, and then clears the KV cache and STM.
The second segment reuses the address with the updated mapping.
After the second consolidation and reset, its held-out query is presented without demonstrations or context-derived STM, so only retained LTM can provide the function parameters sampled for that episode.
Direct LTM conditioning can affect both slot selection while writing the second-segment update and memory traversal during the final query; the present task measures their combined effect.
There is no direct supervision on routes, slots, context tokens, or consolidation boundaries.

\subsection{Mismatched-experience intervention}\label{53-mismatched-experience-intervention}

The mismatched control substitutes a donor experience from the same rule family but with different function parameters, then replaces its address token with the recipient's.
Format, family, and addressing cue are preserved while memory content changes.
Dependence on episode-specific LTM should therefore appear as a collapse in recall.

\section{Experimental Setup}\label{6-experimental-setup}

\subsection{Model and optimization}\label{61-model-and-optimization}

The 29.95M-parameter model has 12 Transformer layers, hidden dimension 384, and a 128-token sliding-attention window.
PMNet uses four hierarchy blocks with branching factor four, 32-dimensional memory, and a 64-dimensional Consolidator hidden layer.

We use AdamW with learning rate \(5\times10^{-4}\), global batch size 256, 100K procedural memory episodes per epoch, and at most 60 epochs.
Validation and test each contain 1K memory episodes.
Five consolidation-training seeds \(\{42,43,44,45,46\}\) use paired data streams and one fixed held-out test stream.
Full architecture, optimizer, software, and seed settings are listed in Appendix~\ref{appendix:hyperparameters}.

\subsection{Parameter-isolation conditions}\label{62-parameter-isolation-conditions}

\begin{table}[t]
\centering
\caption{\textbf{Parameter-isolation conditions.} The routing-off condition changes the forward path but has the same trainable parameter count as the standard Consolidator-only condition.}
\label{tab:parameter-isolation-conditions}
\scriptsize
\setlength{\tabcolsep}{2.5pt}
\begin{tabular}{@{}lllr@{}}
\toprule
Condition & Consolidation & Trainable subset & Parameters \\
\midrule
Learned full & Learned & Entire model & 29.95M (100\%) \\
Identity full & Raw STM accumulation & All except Consolidator & 29.94M (99.959\%) \\
Consolidator only & Learned & Consolidator only & 12.35K (0.041\%) \\
Consolidator only, routing off & Learned & Consolidator only & 12.35K (0.041\%) \\
Memory + Consolidator & Learned & Memory read/write/routing + Consolidator & 1.526M (5.095\%) \\
Learned full, dual objective & Learned & Entire model & 29.95M (100\%) \\
\bottomrule
\end{tabular}
\end{table}

\texttt{Learned full} and \texttt{identity full} are independently optimized; their difference is not a same-checkpoint causal estimate.
The direct mechanism test is \texttt{Consolidator only}, which uses direct LTM-conditioned routing: every other parameter is frozen, and the same trained checkpoint is evaluated with either learned or forced-identity consolidation.
The \texttt{Consolidator only, routing off} condition uses the same parameter isolation but removes the direct LTM term in Equation~\eqref{eq:ltm-conditioned-routing}; both variants retain learned consolidation, the fixed router, and the LTM read path, and each trains its own Consolidator from the same STM-pretraining initialization.

\subsection{Metrics and statistics}\label{63-metrics-and-statistics}

Updated-mapping LTM recall is accuracy on the final second-segment query after both demonstration contexts have been consolidated and the KV cache and STM have been cleared.
Second-segment immediate STM recall evaluates the updated mapping before the second consolidation; mean immediate STM recall averages the corresponding pre-consolidation queries across both segments.
The fresh-LTM control removes persistent memory, whereas the mismatched-LTM control supplies an experience with incorrect function parameters but the correct address and rule family.

We report mean \(\pm\) sample SD over five seeds and use paired differences for comparisons.
Confidence intervals and \(t\)-tests are descriptive because \(n=5\).
All main consolidation-training seeds share one STM-pretraining checkpoint, so their variation reflects consolidation optimization and data streams conditional on that learned STM representation.

\section{Results}\label{7-results}

\subsection{A learned Consolidator enables persistent updates from frozen STM}\label{71-a-separately-learned-consolidator-updates-frozen-stm}

The central intervention freezes every STM-pretrained component that forms, routes, and reads STM and trains only the 12.35K-parameter slot transform.

\begin{table}[t]
\centering
\caption{\textbf{Same-address update under the Consolidator-only intervention.} Both columns evaluate the same trained checkpoint.}
\label{tab:same-address-update}
\small
\begin{tabular}{@{}lrr@{}}
\toprule
State queried after consolidation & Learned Consolidator & Forced identity \\
\midrule
Initial mapping after first consolidation & 50.50 \(\pm\) 3.42 & \textbf{86.86 \(\pm\) 0.05} \\
Updated mapping after second consolidation & \textbf{87.02 \(\pm\) 1.76} & 18.32 \(\pm\) 0.04 \\
\bottomrule
\end{tabular}
\end{table}

\begin{figure}[t]
  \centering
  \begin{subfigure}[t]{0.32\textwidth}
    \centering
    \includegraphics[width=\linewidth]{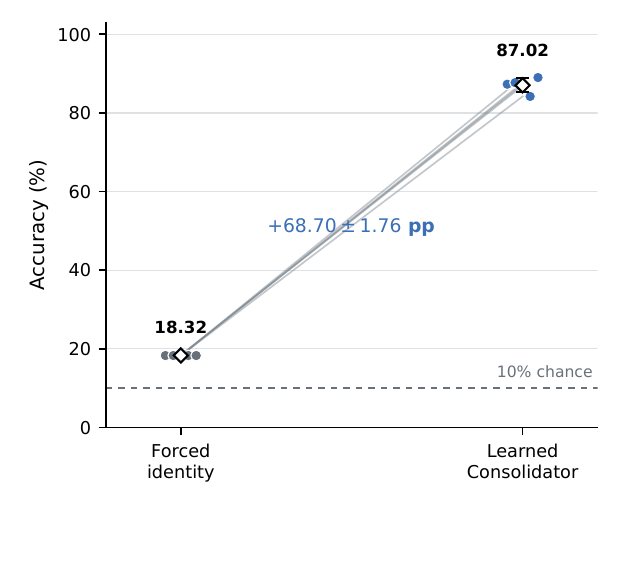}
    \caption{Same-checkpoint boundary intervention.}
    \label{fig:consolidator-boundary}
  \end{subfigure}\hfill
  \begin{subfigure}[t]{0.32\textwidth}
    \centering
    \includegraphics[width=\linewidth]{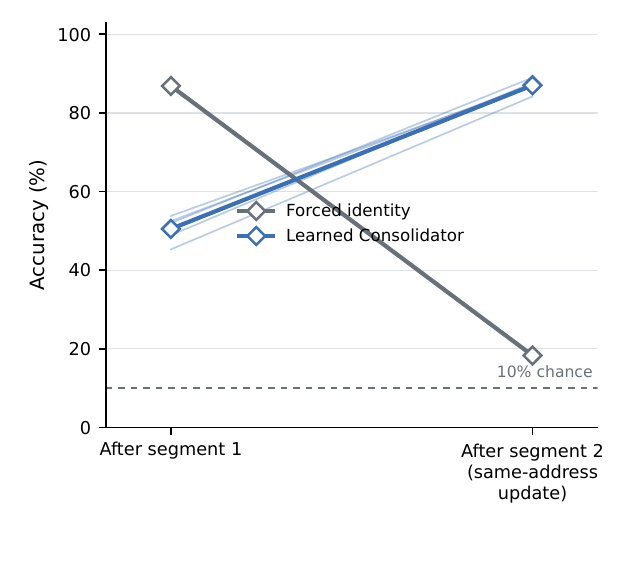}
    \caption{Identity carries; the learned transform updates.}
    \label{fig:consolidator-update}
  \end{subfigure}\hfill
  \begin{subfigure}[t]{0.32\textwidth}
    \centering
    \includegraphics[width=\linewidth]{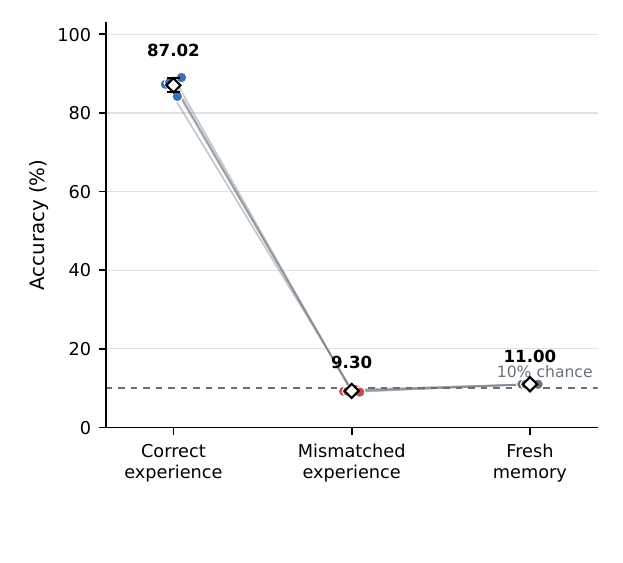}
    \caption{Recall depends on latent content.}
    \label{fig:consolidator-content}
  \end{subfigure}
  \caption{\textbf{Consolidator-only intervention.}
  Paired evaluations compare learned and forced-identity consolidation, the first and second same-address writes, and correct, mismatched, and fresh memory.}
  \label{fig:consolidator-intervention}
\end{figure}

Table~\ref{tab:same-address-update} and Figure~\ref{fig:consolidator-intervention} show that identity accumulation transfers the initial mapping but fails after the second same-address write.
The learned transform instead reaches \(87.02\pm1.76\%\) updated-mapping LTM recall, a same-checkpoint gain of \(68.70\pm1.76\) pp over forced identity while training 0.041\% of the model.
Because the backbone, router, and read/write projections are frozen, the gain must pass through the learned boundary transform and existing memory interface.
The result also provides independent evidence that the upstream STM is functional: because Consolidator receives only routed STM, its slot-local transform could not recover the rule instantiated in the current memory episode unless that STM already encoded the relevant information.
The lower recall of the initial mapping after the first consolidation reflects supervision only on the final updated mapping, not a claim about general retention.

\subsection{Consolidated LTM is an access state, not only stored content}\label{72-consolidated-ltm-conditions-future-access}

\begin{figure}[t]
  \centering
  \begin{subfigure}[t]{0.35\textwidth}
    \centering
    \includegraphics[width=\linewidth]{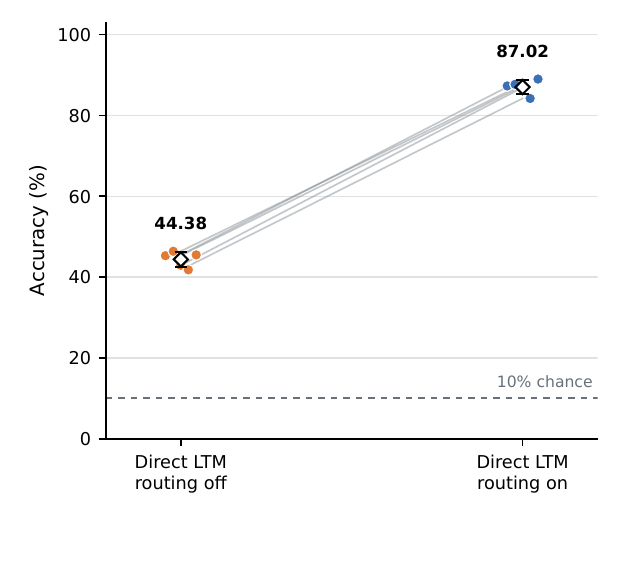}
    \caption{Updated-mapping LTM recall.}
    \label{fig:ltm-routing-recall}
  \end{subfigure}
  \begin{subfigure}[t]{0.35\textwidth}
    \centering
    \includegraphics[width=\linewidth]{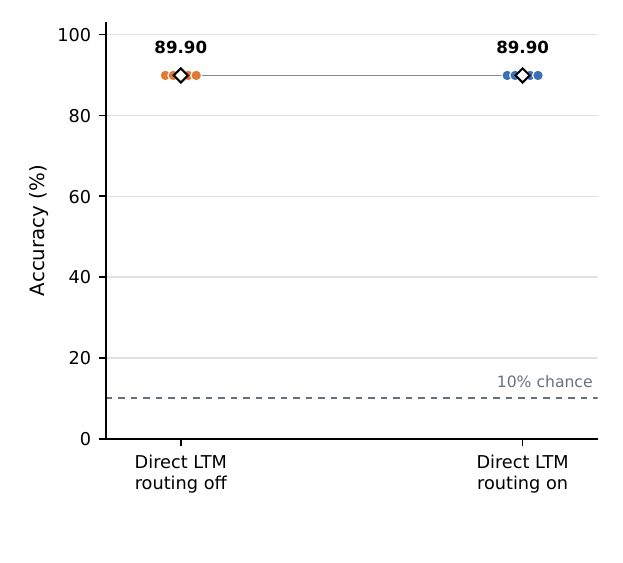}
    \caption{Segment-2 immediate STM recall.}
    \label{fig:ltm-routing-stm}
  \end{subfigure}
  \caption{\textbf{Direct LTM-conditioned routing.}
  Lines pair corresponding consolidation-training seeds initialized from the same STM-pretraining checkpoint.
  Direct routing substantially improves updated-mapping LTM recall (a), while immediate STM recall of the second mapping remains exactly matched (b).
  The off condition retains learned LTM and all read paths; only direct same-level LTM conditioning of the router is removed.}
  \label{fig:ltm-routing}
\end{figure}

Post-reset recall alone does not show that retained LTM affects memory access.
We therefore compare two Consolidator-only conditions that retain learned LTM and all read paths but differ in whether the direct LTM term in Equation~\eqref{eq:ltm-conditioned-routing} is present; each condition trains its own Consolidator.
Table~\ref{tab:direct-ltm-routing} and Figure~\ref{fig:ltm-routing} show that direct routing raises updated-mapping LTM recall from $44.38\pm1.94\%$ to $87.02\pm1.76\%$, a paired gain of $42.64\pm1.10$ pp (95\% CI $[41.27,44.01]$, $p=1.07\times10^{-7}$), while immediate STM recall remains exactly 89.90\% in both conditions.

\begin{table}[t]
\centering
\caption{\textbf{Direct LTM-conditioned routing in the Consolidator-only setting.} On and off runs share a byte-identical STM-pretraining initialization and paired consolidation-training seeds.}
\label{tab:direct-ltm-routing}
\scriptsize
\setlength{\tabcolsep}{3.2pt}
\begin{tabular}{@{}lrrrrrr@{}}
\toprule
Direct LTM routing & Learned LTM & Identity LTM & Learned - identity & Segment-2 STM & Mismatched LTM & Fresh LTM \\
\midrule
Off & 44.38 \(\pm\) 1.94 & 22.98 \(\pm\) 0.04 & +21.40 \(\pm\) 1.91 & \textbf{89.90 \(\pm\) 0.00} & 9.92 \(\pm\) 0.69 & 11.00 \(\pm\) 0.00 \\
On & \textbf{87.02 \(\pm\) 1.76} & 18.32 \(\pm\) 0.04 & \textbf{+68.70 \(\pm\) 1.76} & \textbf{89.90 \(\pm\) 0.00} & 9.30 \(\pm\) 0.24 & 11.00 \(\pm\) 0.00 \\
\bottomrule
\end{tabular}
\end{table}

Without direct routing, learned consolidation still exceeds forced identity by $21.40\pm1.91$ pp, showing that LTM remains useful through the read path; direct routing provides the larger additional gain.
Thus, consolidated LTM serves as both retrievable content and an access state that guides subsequent slot selection; Appendix~\ref{appendix:routing-scope} reports per-seed and rule-family results.

\subsection{Recall requires the correct experience}\label{73-recall-requires-the-correct-latent-experience}

Figure~\ref{fig:consolidator-content} and the routing-on row of Table~\ref{tab:direct-ltm-routing} show that updated-mapping LTM recall from the same checkpoint falls from 87.02\% with the correct experience to \(9.30\pm0.24\%\) with a mismatched experience and 11.00\% with fresh memory.
The mismatch preserves address and rule family, while function parameters vary across memory episodes.
Updated-mapping recall therefore depends on the consolidated content rather than a fixed address association or the mere presence of memory.

\subsection{Broader parameter-isolation checks}\label{74-core-parameter-isolation-ablations}

Under the parameter-isolation conditions summarized in Table~\ref{tab:parameter-isolation-conditions}, learned full reaches 93.70\% and independently trained identity full reaches 91.34\%; their \(+2.36\pm3.30\) pp difference is not statistically resolved.
We therefore do not claim that learned consolidation dominates a fully plastic identity system on this task.
Training only the memory path and Consolidator reaches \(90.70\pm0.51\%\), showing that adaptation can be concentrated in the memory subsystem; Tables~\ref{tab:core-parameter-isolation} and~\ref{tab:core-results-per-seed} report the aggregate and per-seed results in Appendix~\ref{appendix:core-seeds}.

\subsection{Pre-consolidation STM and post-reset LTM recall can coexist}\label{75-immediate-stm-and-final-ltm-can-coexist}

\begin{figure}[t]
  \centering
  \begin{subfigure}[t]{0.35\textwidth}
    \centering
    \includegraphics[width=\linewidth]{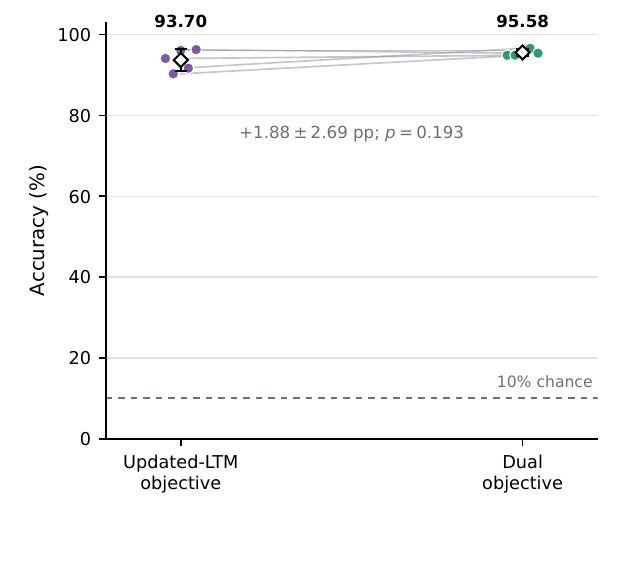}
    \caption{Updated-mapping LTM recall.}
    \label{fig:dual-objective-ltm}
  \end{subfigure}
  \begin{subfigure}[t]{0.35\textwidth}
    \centering
    \includegraphics[width=\linewidth]{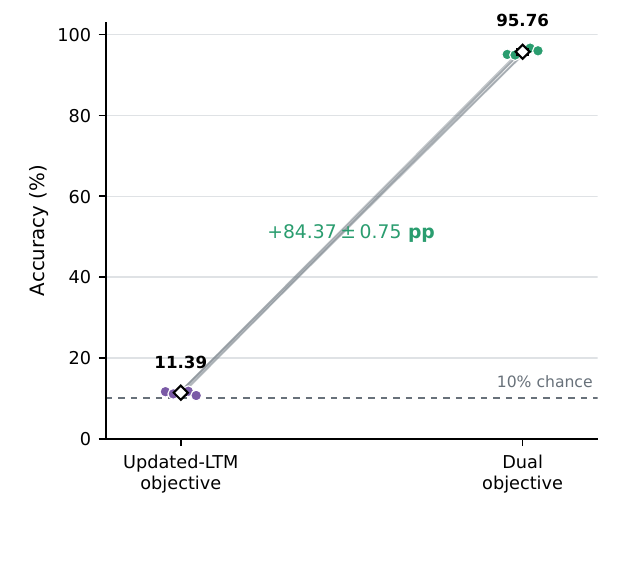}
    \caption{Mean immediate STM recall.}
    \label{fig:dual-objective-stm}
  \end{subfigure}
  \caption{\textbf{Pre-consolidation and post-reset recall under a dual objective.}
  Adding immediate-STM supervision restores mean pre-consolidation recall across both segments (b) without reducing updated-mapping LTM recall (a).
  The two metrics are evaluated on separate trajectories using one shared parameter set.}
  \label{fig:dual-objective}
\end{figure}

Adding pre-consolidation STM supervision raises mean immediate recall across both segments from \(11.39\pm0.47\%\) to \(95.76\pm0.72\%\), while updated-mapping LTM recall reaches \(95.58\pm0.75\%\) (Figure~\ref{fig:dual-objective}).
The two metrics use separate cache trajectories and forward passes, not concurrent queries in one online trajectory.
The result therefore shows that one parameter set can support both capabilities under a suitable objective, not that both were jointly read in a single pass; Appendix~\ref{appendix:dual-seeds} reports aggregate statistics and per-seed results.

ADD10 is nearly saturated, but the Consolidator-only model also reaches \(74.80\pm3.07\%\) on AFFINE10; the central result is therefore not explained only by the simpler additive family.

\section{Discussion}\label{8-discussion}

The results establish a functional chain: STM pretraining forms episode-specific routed state, Consolidator converts it for persistent revision, and replacing the retained experience removes the recall gain.
Because the upstream memory interface is frozen and Consolidator receives no source tokens, its success implies that STM already contains the relevant information before consolidation.

The identity and routing interventions show that raw persistence and readout alone do not explain the result: the learned boundary supports conflicting revision, while direct routing provides the larger gain despite matched immediate STM recall.
Consolidated LTM therefore functions as both retrievable content and an access state, making its direct connection to the frozen router a task-effective inductive bias for experience-dependent slot selection.

At inference, the fixed Consolidator and router adapt through mutable non-parametric state rather than parameter updates, distinguishing the mechanism from continual fine-tuning and test-time gradient descent.

\section{Limitations}\label{9-limitations}

\textbf{Controlled scope.}
Memory episodes contain two short context segments, one active address, and modular-arithmetic rules.
Both demonstration contexts fit inside the local attention window; the reset isolates persistence but does not test extreme within-segment context, natural language, many competing memories, long horizons, or systems efficiency.

\textbf{Estimation and provenance.}
The five main consolidation-training seeds share one selected STM-pretraining representation, so their variance excludes variation from the first training stage.
Training retains gradients across the two consolidation boundaries, leaving detached or truncated long-horizon training untested.
With \(n=5\), confidence intervals and \(t\)-tests are descriptive.

\textbf{Unisolated design choices.}
Identity is the principal same-checkpoint control, but we do not compare alternative learned overwrite, EMA, linear, or gated recurrent operators.
Boundaries, commit decisions, and eviction are externally specified, and the synthetic task lacks a task-matched external architecture baseline.

\textbf{Persistence semantics.}
LTM is initialized for each memory episode; persistence across unrelated sessions, serialization, and deployment restarts is not evaluated.
STM, LTM, and consolidation denote computational timescales, not a biological model of memory or sleep.

\section{Conclusion}\label{10-conclusion}

We introduced Consolidator, a shared slot-local transform that converts routed STM into persistent LTM without replaying the source tokens.
On a controlled same-address update task, training only its 12.35K parameters while freezing the rest of PMNet yields 87.02\% updated-mapping recall, compared with 18.32\% when the same checkpoint uses identity accumulation; replacing the retained experience removes this gain.
A paired routing ablation further reduces recall from 87.02\% to 44.38\% while leaving immediate STM recall unchanged, showing that consolidated LTM supports later computation both as retrievable content and as an input to slot selection.
These results establish a controlled forward-state adaptation mechanism, not yet a general long-term memory system; detached or truncated long-horizon training, natural language, and scale-up remain open.

\backmatter
\section{Reproducibility}\label{reproducibility-statement}

The implementation will be made publicly available at
\url{https://www.github.com/swgoo/pmnet_consolidator}.
It records full run configurations, starting-checkpoint hashes, trainable parameter counts, per-seed selected checkpoints, and the fixed test stream shared across conditions.
\texttt{modeling\_pmnet.py} contains the cache, routing, Consolidator, and persistent-state update; \texttt{train\_pmnet\_ablation.py} contains the procedural data generator, STM-pretraining stage (Phase 1), interventions, and consolidation-training runner (Phase 2).
Per-seed results and complete hyperparameters are reported in the appendix.

\bibliographystyle{unsrtnat}
\bibliography{ref}
\clearpage
\appendix
\makeatletter
\def\fps@table{htbp}
\def\fps@figure{htbp}
\makeatother
\counterwithin{table}{section}
\counterwithin{figure}{section}
\renewcommand{\thetable}{\thesection\arabic{table}}
\renewcommand{\thefigure}{\thesection\arabic{figure}}
\setcounter{secnumdepth}{3}
\section{Direct LTM Routing by Seed and Rule Family}\label{appendix:routing-scope}

Table~\ref{tab:routing-per-seed} reports the pooled paired routing results for each seed.
Table~\ref{tab:routing-rule-family} then stratifies learned-LTM recall by rule family.
Direct LTM routing improves both families: the simpler ADD10 family approaches saturation, while AFFINE10 retains a paired gain of \(38.56\pm1.29\) pp.
The overall routing effect is therefore not attributable only to the additive rules.

\begin{table}[htbp]
\centering
\caption{\textbf{Per-seed direct LTM-routing ablation in the Consolidator-only setting.} Routing-on and routing-off runs start from the byte-identical Phase-1 checkpoint and use paired Phase-2 seeds, but each trains its own Consolidator. \emph{Off learned LTM} is updated-mapping recall after both consolidations and the final reset with direct same-level LTM routing disabled; \emph{Off identity LTM} reevaluates that same routing-off checkpoint using forced identity accumulation, and \emph{Learned--identity} is their same-checkpoint difference. Segment-2 STM, mismatched LTM, and fresh LTM are also measured in the routing-off condition. \emph{On--Off} pairs learned-LTM recall from the standard routing-on run with learned-LTM recall from the routing-off run. Both routing conditions retain the fixed router and all LTM read paths. All values are percentages on the shared fixed test stream.}
\label{tab:routing-per-seed}
\scriptsize
\setlength{\tabcolsep}{2.2pt}
\begin{tabular}{@{}rrrrrrrr@{}}
\toprule
Seed & Off learned LTM & Off identity LTM & Learned - identity & Segment-2 STM & Mismatched LTM & Fresh LTM & On - Off \\
\midrule
42 & 45.30 & 23.00 & +22.30 & 89.90 & 9.30 & 11.00 & +42.00 \\
43 & 46.40 & 23.00 & +23.40 & 89.90 & 10.40 & 11.00 & +41.30 \\
44 & 42.90 & 23.00 & +19.90 & 89.90 & 10.70 & 11.00 & +44.00 \\
45 & 41.80 & 22.90 & +18.90 & 89.90 & 9.10 & 11.00 & +42.40 \\
46 & 45.50 & 23.00 & +22.50 & 89.90 & 10.10 & 11.00 & +43.50 \\
\midrule
Mean & 44.38 & 22.98 & +21.40 & 89.90 & 9.92 & 11.00 & +42.64 \\
SD & 1.94 & 0.04 & 1.91 & 0.00 & 0.69 & 0.00 & 1.10 \\
\bottomrule
\end{tabular}
\end{table}

\begin{table}[htbp]
\centering
\caption{\textbf{Direct LTM-routing ablation by rule family.} Routing-off and routing-on values are updated-mapping LTM recall from separately trained Consolidator-only runs that share the byte-identical Phase-1 checkpoint and paired Phase-2 seeds. On--Off differences are computed within each seed before aggregation. Values are percentages, reported as mean \(\pm\) sample SD over five seeds on the shared fixed test stream.}
\label{tab:routing-rule-family}
\small
\setlength{\tabcolsep}{7pt}
\begin{tabular}{@{}lrrr@{}}
\toprule
Rule family & Routing off & Routing on & Paired On--Off \\
\midrule
ADD10 & 52.52 \(\pm\) 1.37 & \textbf{99.01 \(\pm\) 0.60} & \textbf{+46.49 \(\pm\) 1.23} \\
AFFINE10 & 36.24 \(\pm\) 2.61 & \textbf{74.80 \(\pm\) 3.07} & \textbf{+38.56 \(\pm\) 1.29} \\
\bottomrule
\end{tabular}
\end{table}

\clearpage
\section{Exact Memory-Episode Procedure}\label{appendix:memory-episode-procedure}

\subsection{Pseudocode}\label{a1-pseudocode}

\begin{verbatim}
sample family f in {ADD10, AFFINE10}
sample address a from four split tokens
sample first-segment parameters theta_1
sample second-segment parameters theta_2 != theta_1

for segment d in {1, 2}:
    sample 8 distinct demonstration inputs
    sample 1 held-out query input
    if d == 2:
        require f_theta_1(query) != f_theta_2(query)
    form 40-token demonstration context
    process context with routed STM writes
    consolidate occupied STM groups into LTM
    clear KV and STM; retain LTM

present the second-segment held-out query without demonstrations
apply loss only to the answer token
\end{verbatim}

For the dual objective, a separate immediate-STM trajectory reprocesses each segment's context and adds its pre-consolidation query loss to the updated-mapping LTM loss.

\clearpage
\section{Per-Seed Core Results}\label{appendix:core-seeds}

\begin{table}[htbp]
\centering
\caption{\textbf{Core parameter-isolation ablations.} Percent accuracy, mean \(\pm\) SD over five seeds; LTM recall follows the final reset.}
\label{tab:core-parameter-isolation}
\scriptsize
\setlength{\tabcolsep}{3pt}
\begin{tabular}{@{}lrrrr@{}}
\toprule
Condition & Updated LTM & Fresh & Mismatched & Segment-2 STM \\
\midrule
Learned full & \textbf{93.70 \(\pm\) 2.66} & 10.84 \(\pm\) 0.54 & 7.70 \(\pm\) 0.98 & 11.94 \(\pm\) 0.67 \\
Identity full & \textbf{91.34 \(\pm\) 2.90} & 11.12 \(\pm\) 0.28 & 7.92 \(\pm\) 1.02 & 18.16 \(\pm\) 4.34 \\
Consolidator only & \textbf{87.02 \(\pm\) 1.76} & 11.00 \(\pm\) 0.00 & 9.30 \(\pm\) 0.24 & \textbf{89.90 \(\pm\) 0.00} \\
Memory + Consolidator & \textbf{90.70 \(\pm\) 0.51} & 10.30 \(\pm\) 0.29 & 9.44 \(\pm\) 0.31 & 14.88 \(\pm\) 3.24 \\
\bottomrule
\end{tabular}
\end{table}

\begin{table}[htbp]
\centering
\caption{\textbf{Fixed-test-stream results for all 20 core runs.} ``Alternative mode'' is a same-checkpoint diagnostic that changes only the consolidation operator at evaluation: learned conditions use forced raw-identity accumulation, whereas identity full uses its frozen, identity-initialized Consolidator path. The alternative mode is not independently trained.}
\label{tab:core-results-per-seed}
\scriptsize
\setlength{\tabcolsep}{2.5pt}
\begin{tabular}{@{}lrrrrrr@{}}
\toprule
Condition & Seed & Primary recall & Alternative mode & Fresh & Segment-2 STM & Mismatched \\
\midrule
Learned full & 42 & 94.10 & 53.20 & 10.90 & 11.50 & 7.60 \\
Learned full & 43 & 90.30 & 87.90 & 10.80 & 11.50 & 9.30 \\
Learned full & 44 & 96.10 & 82.90 & 11.00 & 12.20 & 6.90 \\
Learned full & 45 & 91.70 & 88.50 & 11.50 & 13.00 & 7.80 \\
Learned full & 46 & 96.30 & 75.10 & 10.00 & 11.50 & 6.90 \\
Identity full & 42 & 96.10 & 96.10 & 11.40 & 12.60 & 7.10 \\
Identity full & 43 & 88.20 & 88.10 & 11.10 & 20.20 & 9.10 \\
Identity full & 44 & 90.70 & 90.70 & 10.80 & 23.70 & 8.70 \\
Identity full & 45 & 91.20 & 91.10 & 10.90 & 19.10 & 6.70 \\
Identity full & 46 & 90.50 & 90.20 & 11.40 & 15.20 & 8.00 \\
Consolidator only & 42 & 87.30 & 18.30 & 11.00 & 89.90 & 9.20 \\
Consolidator only & 43 & 87.70 & 18.30 & 11.00 & 89.90 & 9.20 \\
Consolidator only & 44 & 86.90 & 18.40 & 11.00 & 89.90 & 9.50 \\
Consolidator only & 45 & 84.20 & 18.30 & 11.00 & 89.90 & 9.60 \\
Consolidator only & 46 & 89.00 & 18.30 & 11.00 & 89.90 & 9.00 \\
Memory + Consolidator & 42 & 91.60 & 85.00 & 10.50 & 10.70 & 9.20 \\
Memory + Consolidator & 43 & 90.60 & 87.60 & 10.10 & 16.70 & 9.10 \\
Memory + Consolidator & 44 & 90.40 & 72.70 & 10.40 & 18.90 & 9.40 \\
Memory + Consolidator & 45 & 90.40 & 89.30 & 9.90 & 12.70 & 9.70 \\
Memory + Consolidator & 46 & 90.50 & 82.10 & 10.60 & 15.40 & 9.80 \\
\bottomrule
\end{tabular}
\end{table}

\clearpage
\section{Per-Seed Dual-Objective Results}\label{appendix:dual-seeds}

This experiment asks whether the low immediate-STM recall observed when training only for the final post-reset query reflects an architectural incompatibility between STM and LTM, or simply the absence of direct STM supervision.
Both conditions use the fully trainable model, share the same Phase-1 STM-pretraining checkpoint, and pair the Phase-2 seeds and data streams.
They differ in their training objective and checkpoint-selection metric, so this comparison is between independently optimized objectives rather than a same-checkpoint intervention.

The \emph{Updated LTM only} condition optimizes the final second-segment query after both consolidations and the final KV/STM reset.
The \emph{Updated LTM + STM} condition adds the mean loss of two auxiliary pre-consolidation queries, one for each segment.
These immediate-STM queries are evaluated on separate auxiliary trajectories; they share model parameters with the persistent-memory trajectory but do not interrupt or supply information to it.
In the tables, \emph{Updated LTM} is final updated-mapping recall, \emph{Mean STM} averages immediate recall across the two segments, and \emph{Segment-2 STM} reports immediate recall of the updated mapping alone.

\begin{table}[htbp]
\centering
\caption{\textbf{Aggregate dual-objective experiment.} Paired runs share seeds and STM-pretraining provenance; the consolidation-training objective and validation monitor differ.}
\label{tab:dual-objective-aggregate}
\scriptsize
\setlength{\tabcolsep}{3pt}
\begin{tabular}{@{}lrrrrr@{}}
\toprule
Training objective & Updated LTM & Mean STM & Segment-2 STM & Mismatched & Fresh \\
\midrule
Updated LTM only & 93.70 \(\pm\) 2.66 & 11.39 \(\pm\) 0.47 & 11.94 \(\pm\) 0.67 & 7.70 \(\pm\) 0.98 & 10.84 \(\pm\) 0.54 \\
Updated LTM + STM & \textbf{95.58 \(\pm\) 0.75} & \textbf{95.76 \(\pm\) 0.72} & \textbf{95.86 \(\pm\) 0.91} & 8.54 \(\pm\) 0.30 & 10.96 \(\pm\) 0.67 \\
Paired change & +1.88 \(\pm\) 2.69 & \textbf{+84.37 \(\pm\) 0.75} & \textbf{+83.92 \(\pm\) 0.59} & --- & --- \\
95\% CI & {[}-1.46, +5.22{]} & \textbf{{[}+83.44, +85.30{]}} & \textbf{{[}+83.19, +84.65{]}} & --- & --- \\
Paired \(p\) & 0.193 & \textbf{\(1.52\times10^{-9}\)} & \textbf{\(5.83\times10^{-10}\)} & --- & --- \\
\bottomrule
\end{tabular}
\end{table}

Adding direct STM supervision raises mean immediate recall by \(84.37\pm0.75\) pp while retaining \(95.58\pm0.75\%\) updated-mapping LTM recall.
The \(+1.88\pm2.69\) pp change in updated LTM recall is not statistically resolved; the result therefore supports coexistence of the two capabilities under one parameter set, not an improvement in LTM attributable to the auxiliary objective.
The paired confidence intervals and \(t\)-tests are descriptive because \(n=5\).

\begin{table}[htbp]
\centering
\caption{\textbf{Per-seed dual-objective results.}}
\label{tab:dual-objective-per-seed}
\scriptsize
\setlength{\tabcolsep}{3pt}
\begin{tabular}{@{}rrrrrrr@{}}
\toprule
Seed & Best epoch & Validation dual & Test dual & Updated LTM & Mean STM & Forced identity \\
\midrule
42 & 10 & 95.07 & 95.00 & 94.90 & 95.10 & 84.60 \\
43 & 14 & 95.42 & 94.92 & 94.90 & 94.95 & 72.30 \\
44 & 20 & 95.90 & 96.10 & 96.10 & 96.10 & 89.20 \\
45 & 22 & 96.20 & 96.63 & 96.60 & 96.65 & 67.30 \\
46 & 16 & 95.10 & 95.70 & 95.40 & 96.00 & 71.70 \\
\bottomrule
\end{tabular}
\end{table}

For the per-seed table, \emph{Validation dual} and \emph{Test dual} are the arithmetic means of updated-mapping LTM recall and mean immediate-STM recall on their respective splits.
\emph{Forced identity} reevaluates the selected dual-objective checkpoint with raw identity accumulation in place of the learned Consolidator; it is a same-checkpoint diagnostic, not a separately trained identity condition.

\clearpage
\clearpage
\section{Hyperparameters and Architecture}\label{appendix:hyperparameters}

\begin{table}[htbp]
\centering
\caption{\textbf{Model, optimization, and hardware settings.}}
\label{tab:hyperparameters}
\scriptsize
\begin{tabularx}{\textwidth}{@{}>{\raggedright\arraybackslash}p{0.25\textwidth}X@{}}
\toprule
Category & Setting \\
\midrule
Backbone initialization & PMNet copy-task checkpoint, followed by same-segment STM rule induction (Phase 1) \\
Total parameters & 29.95M \\
Transformer & 12 layers; hidden size 384; FFN size 1,024; 12 query and 12 KV heads \\
Local attention & Sliding window 128; attention dropout 0.1; RMSNorm \(\epsilon=10^{-6}\) \\
Memory hierarchy & Four blocks; branch factor 4; 85 routing groups; 340 candidate slot vectors \\
Memory features & Phase dimension 32; four read heads; writes at layers 0, 3, 6, and 9 \\
Consolidator & Intermediate size 64; SiLU gate; 12.35K shared parameters; identity phase initialization \\
Optimizer & AdamW; \(\beta=(0.9,0.95)\); learning rate \(5\times10^{-4}\); weight decay 0.1 \\
Schedule & 100 warmup steps, then cosine decay \\
Gradient clipping & Global norm 1.0 \\
Precision and batch & \texttt{bf16-mixed}; global batch size 256; one device \\
Hardware & One NVIDIA RTX 4090; CUDA 13.0; FlashAttention 2 \\
Software & Python 3.12.3; PyTorch 2.10.0+cu130; Transformers 5.14.1; Lightning 2.6.5 \\
Training budget & 100K memory episodes/epoch; at most 60 epochs; patience 6 \\
Evaluation & 1K validation and 1K fixed test memory episodes \\
Phase-2 seeds (consolidation) & 42, 43, 44, 45, 46 \\
Temporal graph & \texttt{detach\_long\_term\_between\_sleeps=False} for reported adaptation runs \\
\bottomrule
\end{tabularx}
\end{table}

\end{document}